\documentclass{article}
\usepackage{ijcai23}

\usepackage{times}
\usepackage{soul}
\usepackage{url}
\usepackage[hidelinks]{hyperref}
\usepackage[utf8]{inputenc}
\usepackage[small]{caption}
\usepackage{graphicx}
\usepackage{amsmath}
\usepackage{amsthm}
\usepackage{pgf-pie}
\usepackage{booktabs}
\usepackage{algorithm}
\usepackage{algorithmic}
\usepackage{graphicx} 
\usepackage[switch]{lineno}

\title{What's meant by explainable model: A Scoping Review }

\author{
Mallika Mainali$^1$
\and
Rosina O Weber$^1$$^2$
\affiliations
$^1$Information Science, Drexel University\\
$^2$Computer Science, Drexel University\\
\emails
\{mm5579, rosina\}@drexel.edu
}

\begin{document}

\maketitle

\begin{abstract}

%We often see the term \textit{explainable} in the titles of papers that describe AI applications. However, it is uncertain what \textit{explainable} means when AI models are referred to as explainable. Upon examination, we noticed several papers referring to their models as \textit{explainable} when all they had done was to implement popular feature attribution libraries such as LIME or SHAP for a sample of instances. Moreover, the papers we examined seemed to assume that the problem of explainability in machine learning was solved. We examined the articles describing AI applications and analyzed the number of applications that employed an evaluation method for the explanations. Out of 187 studies included in the review, only 35 (18.72\%) used an evaluation approach to assess the quality of their XAI method, out of which 16 studies (8.56\%) utilized the evaluation approach for proposed new explanation methods. We conclude that most models that are presented as explainable do not perform an evaluation of their XAI methods.
%
We often see the term \textit{explainable} in the titles of papers that describe applications based on artificial intelligence (AI). However, the literature in explainable artificial intelligence (XAI) indicates that explanations in XAI are application- and domain-specific, hence requiring evaluation whenever they are employed to explain a model that makes decisions for a specific application problem. Additionally, the literature reveals that the performance of post-hoc methods, particularly feature attribution methods, varies substantially hinting that they do not represent a solution to AI explainability. Therefore, when using XAI methods, the quality and suitability of their information outputs should be evaluated within the specific application. For these reasons, we used a scoping review methodology to investigate papers that apply AI models and adopt methods to generate post-hoc explanations while referring to said models as \textit{explainable}. This paper investigates whether the term \textit{explainable model} is adopted by authors under the assumption that incorporating a post-hoc XAI method suffices to characterize a model as \textit{explainable}. To inspect this problem, our review analyzes whether these papers conducted evaluations. We found that 81\% of the application papers that refer to their approaches as an \textit{explainable model} do not conduct any form of evaluation on the XAI method they used.

\end{abstract}

\section{Introduction}
Methods that attempt to produce information in lieu of explanatory value to make AI methods more transparent are very popular. As of April 2023, Google Scholar has reported about 140,000 papers that contain the term \textit{XAI} in their title \cite{GoogleScholar_XAI}. Due to this popularity, XAI has been the topic of multiple workshops at AI conferences such as AAAI \cite{aaai2022} and IJCAI \cite{ijcai2021,ijcai2022} in the last few years. The papers demonstrating XAI methods describe a variety of topics related to how AI methods can explain their decisions. XAI methods vary in the type of information content they produce as output. The categories of information outputs are feature attributions (\textit{e.g.}, SHAP \cite{lundberg2017unified}, LIME \cite{ribeiro2016why}, DeepLIFT \cite{shrikumar}), counterfactuals (\textit{\textit{e.g.}}, DiCE \cite{Mothilal}, MACE \cite{yang2022mace}), instance attributions (\textit{e.g.}, influence functions \cite{koh2017understanding}, representer points \cite{yeh2018representer}, HYDRA \cite{hydra}), example-based (\textit{e.g.}, \cite{cbr-nugent,Keane_2019}), and method-specific outputs such as decision tree paths (\textit{e.g.}, \cite{izza2022tackling}). 

Among these multiple XAI methods, SHAP \cite{lundberg2017unified} and LIME \cite{ribeiro2016why} have easy-to-use libraries that help them be quite popular. As of May 2023, the papers proposing SHAP and LIME have been cited, respectively, 12,740 \cite{GoogleScholar_Shap} and 12,504 \cite{GoogleScholar_Lime} times, while papers referencing instance attribution like influence functions have only been cited 2,193 times \cite{GoogleScholar_Influence_functions}. Although these feature attribution methods are popular and easy to use, using these libraries under the assumption that the problem of explaining a model's decision is solved is not aligned with the body of literature that demonstrates their limitations and variability, and with those that prescribe domain- and application-specific evaluations \cite{lin2020see,VANDERWAA2021103404,metrics_for_evaluating_rosenfeld,zhou2021feature,rong2022humancentered,Marques-Silva_Ignatiev_2022}. Using these libraries with the claim of explainability without proper evaluation can be misleading because these feature attribution methods have been found to select misleading non-correlating features \cite{zhou2021feature}.

Despite the popularity of many XAI methods, the literature often mentions criticisms of these methods (\textit{e.g.}, \cite{adebayo2018sanity,zhou2021feature,Marques-Silva_Ignatiev_2022,huang2023inadequacy}). Said criticisms justify these methods be carefully evaluated to determine their suitability to specific application contexts, particularly because it is consensual that explanations are application-specific and contextual (\textit{e.g.}, dependent on user, domain, and task \cite{8012316,doshivelez2017rigorous,doi:10.1126/scirobotics.aay7120,mueller2019explanation,BARREDOARRIETA202082,zhou2021}).

Another aspect to consider is that the output produced by feature attribution methods is the contributions of the features in each decision. This is considered important information with explanatory value but it is only one of many categories of information outputs that can be produced. They are not sufficient to describe the complete decision making process of a model. This limited scope combined with their variability in quality is yet another reason to include application-specific evaluations as many domains may be considered high-stakes such as healthcare, finance, privacy and security (\textit{e.g.}, \cite{rudin2019stop,Marques-Silva_Ignatiev_2022,rudin2022interpretable}. 

When examining publications using the term \textit{explainable} to describe machine learning (ML) methods, and particularly referring to such methods as \textit{explainable models}, we noticed that many of those papers implemented one of the popular XAI libraries without any evaluation. By not evaluating these XAI methods, these works are sending a misleading message that these XAI methods have solved the explainability problem and simply generating feature attributions makes their model \textit{explainable}. Based on the literature as mentioned above, there needs to be a scientific evaluation of the applications that use XAI methods to confirm if those applications produce reliable explanations for the specific application and domain where the models are applied. This is because existing XAI methods can often be wrong, and can possibly even be dangerous if they are wrong when used in safety-critical applications \cite{rudin2019stop,rudin2022interpretable}. If these methods are being used to explain the decisions of AI models, the explanations need to be evaluated to ensure they are reliable in all applications, particularly high-stakes and safety-critical applications. 

To get a scientific perspective on how often the XAI methods are evaluated when used to make a model explainable, in this paper, we employed a scoping literature review methodology. With the scoping review, we aim to gather scientific evidence on the use of the term \textit{explainable model} with and without the presence of evaluation of XAI methods. To achieve this, this study will address the following research question (RQ): 

%Research Question (RQ): How extensive is the practice of referring to a model as explainable simply because SHAP or LIME libraries were used to provide feature attributions without evaluating the XAI method used? %
\paragraph{RQ:} How extensive is the practice of referring to a model as explainable simply because methods to provide feature attributions were used without evaluation of the explanations generated by the method?

In our study, we categorize an XAI method as evaluated if the authors of these papers conduct any qualitative or quantitative scientific experiment (\textit{e.g.}, case studies, computational evaluation using cross-validation, fidelity test) to assess the quality of feature attributions produced by adopted AI methods. The next section of this paper describes the background of XAI methods and their evaluation approaches. Section 2.1 describes the advancements and limitations of feature attribution methods such as SHAP \cite{lundberg2017unified} and LIME \cite{ribeiro2016why}. Section 2.2 discusses the state of evaluation approaches for \textit{post hoc} explanation methods. We present the the scoping review methodology in Section 3. In section 4, we show an analysis of the results and conclude with a discussion on the state of evaluation methods used for so-called \textit{explainable models}. 
%In this study, we wanted to get a scientific perspective on how often the XAI methods illustrated in the literature are evaluated. Although it is critical to evaluate XAI methods when using them in application development, our observation has shown that XAI methods illustrated in the scientific literature are often not evaluated. % 

\begin{table*}
    \centering
    \begin{tabular}{lcc}
        \toprule    
        & XAI methods evaluated  & XAI methods not evaluated \\
        \midrule
        New method of explanation proposed& 16& 33\\
        Use of SHAP library& 10& 92\\
        Use of LIME library& 3& 14\\
        Combination of different XAI methods (SHAP and LIME)& 6& 13\\  
        \bottomrule
\end{tabular}
    \caption{Classification of literature based on evaluation of XAI methods}
    \label{tab:booktabs}
\end{table*}
\section{Background}
\subsection{Background of XAI Methods}
With the development of feature-attribution methods like LIME in 2016 \cite{ribeiro2016why} and SHAP in 2017 \cite{lundberg2017unified}, model explainability has become an important aspect in designing black-box models. Recent trends show that the popularity of XAI methods has been growing exponentially over the last 5 years with the saturation in regards to deep learning \cite{angelov}. While these methods do provide some interpretation of a model’s predictions, the explanations generated may not always be informative, or accurate (\textit{e.g.}, \cite{rudin2019stop,rudin2022interpretable,Marques-Silva_Ignatiev_2022}). Multiple studies have pointed out that the use of SHAP or LIME libraries does not necessarily create an explanatory model and have dissected the potential limitations of these libraries \cite{hatwell,almerri}. Since there is no formal definition of feature importance, it is difficult to validate these methods \cite{wilming2021scrutinizing}. Because of these drawbacks, many researchers have been investigating the challenges and limitations of XAI methods, which stem from the explanation generation without the quantification of the explanation \cite{Molnar_2020}. 

Angelov et al. dissected the use of different types of explanations, including SHAP and LIME, to demonstrate that the explanations generated by these models are not always efficient or reliable \cite{angelov}. Similarly, Tan et al. have described the uncertainty of explanations \cite{tan}, and Rengasamy et al. have referenced the instability of LIME by citing its inability to give a global approximation of feature importance \cite{rengasamy}. Additionally, multiple studies have attempted to analyze the explanations of the predictions made by two models to compare their soundness, consistency, and accuracy \cite{slack,alharbi}. For instance, Alharbi et al. demonstrated the inconsistency between the explanations generated using SHAP of two models of the same image and proposed a novel approach to minimize the inconsistency \cite{alharbi}. Similarly, a demonstration by Slack et al. revealed the weakness of \textit{post hoc} explanation techniques by explaining how these methods could be easily fooled with adversarial classifiers \cite{slack}.

Some studies have showcased the need for the evaluation of XAI methods by demonstrating the inefficiency of the explanations generated by these methods after an evaluation with human experts \cite{Rafferty}. Rafferty et al. showed the unreliability of using \textit{post hoc} explanation methods in high-stake decision-making by revealing that the explanations from the three techniques (\textit{i.e.}, SHAP, LIME, and RISE) were unhelpful \cite{Rafferty}. This has led to the development of frameworks for explanations in sensitive fields like healthcare \cite{barda}, but a standardized framework for the evaluation of those explanations is still needed.

\subsection{Evaluation of \textit{post hoc} explanations}
Despite the popularity of \textit{post hoc} explanation methods, there are no standard methods to evaluate them. Some studies have shown the necessity of developing a framework for the evaluation of XAI methods \cite{rawal}, and there have been works that proposed evaluation criteria for the quality of explanations (\textit{e.g.}, \cite{sundararajan2017axiomatic,alavarez2018,cui2019integrative,Montavon2019}). These criteria reiterate the need to evaluate an explanation based on different dimensions such as human comprehensibility, fidelity, accuracy, scalability, and generality \cite{Dai_2022,belle}. Despite the importance of an evaluation metric for XAI methods, there is currently no commonly recognized scale that is widely accepted by XAI researchers \cite{LIU2022102111}. This could be because of the difficulty and the expense needed to quantify the explanations as Zhou et al. have demonstrated that the evaluation of the explanations is a multidisciplinary research area, which makes it difficult to define an implementation of a metric that can be applied to all applications of XAI methods \cite{zhou2021}. Nonetheless, Vu et al. have introduced x-Eval which is an evaluation metric that quantifies feature-based local explanation’s quality \cite{vu2019}, and Cugny et al. have introduced AutoXAI, which provides a framework for automatically selecting an XAI method based on some evaluation metrics \cite{cugny}. These attempts, however, have not been widely adopted. 

\section{Methodology}
In this section, we describe the implementation o the scoping review methodology outlined by Arksey and O'Malley (2005). A scoping review collects and categorizes existing literature and identifies the nature and gaps of current research evidence. Contrary to systematic reviews that seek to answer pre-defined questions from a narrow range of quality-assessed evidence, scoping reviews examine the range of a research topic and identify the gaps within existing literature \cite{Arksey}. The scoping review fits our goal, which is to examine the range of literature that use the term \textit{explainable model} to indicate a model for which a \textit{post hoc} XAI method was applied and to analyze the literature that demonstrates the lack of evaluation. Next, we describe data extraction and collection.

\subsection{Data Extraction}
The initial step in data extraction is to search the literature. Our primary focus was to identify papers describing the use of some model to execute an AI task such as classification in a given domain so that the papers can be characterized as \textit{application papers} rather than \textit{research papers}. It is typically when applying AI models within a specific application that users and consequently model explainability is considered. We had anecdotally observed that many papers were adopting some \textit{post hoc} XAI method to generate feature attributions in lieu of explanation for those applications of models and referring to them as \textit{explainable models}.

We executed the search targeting the period starting in January 2016 and going until December 2022. We started the search on the Scopus database for articles that contained the following keywords: explainable* OR explainability* OR interpretability* OR explainable approach* OR explainable method* OR explainable AI* OR explainable machine learning* OR XAI*. The other inclusion criteria were English language and application papers. To facilitate the search and the analysis, we included papers that referenced either of the two most commonly used feature attribution methods (\textit{e.g.}, SHAP or LIME). This last criterion helped us select the right papers because of their popularity.

\begin{figure}[htp]
\centering
\includegraphics[width=\linewidth]{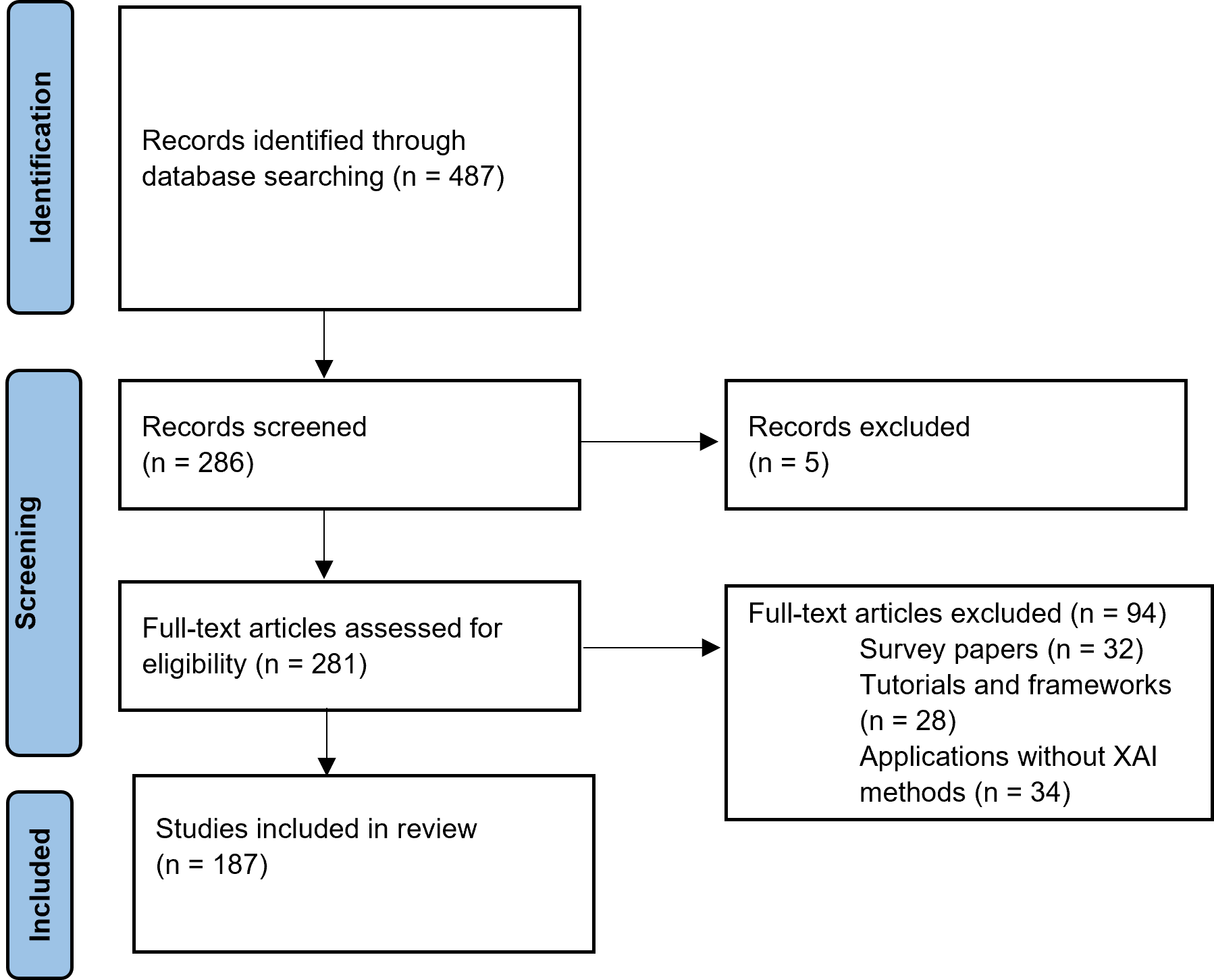}
\caption{PRISMA flow diagram for the selection process of literature for the scoping review of the evaluation of XAI methods in application papers.}
\label{fig:prisma}
\end{figure}

\subsection{Data Collection}
Figure \ref{fig:prisma} shows the steps Identification, Screening, and Inclusion describing how we moved from 487 initial papers to the final 187 papers. 
The screening process was conducted using the PRISMA extension \cite{prisma} for scoping reviews (Figure~\ref{fig:prisma}). The titles and abstracts of the papers were screened to check for eligibility. The initial search in Scopus resulted in 487 articles that described some applications using XAI methods. These 487 articles were further filtered to 286 articles based on open-access availability. After the initial screening, all the data from the included articles were extracted for manual analysis. Out of 286 articles, five articles were excluded because they were not downloadable for data analysis. 281 full-text articles were downloaded and reviewed manually. The process of manual review consisted of this first author examining the full-text to ascertain the following aspects: 1. Whether the paper was indeed applying an AI model representing an application paper; 2. If the title or any other section described their applied model as \textit{explainable} or \textit{interpretable}; 3. If the paper really adopted a \textit{post hoc} XAI method; and 4. If the XAI method was evaluated. 

After this thorough review process, 94 articles were excluded. The reasons for excluding these 94 articles were: 32 articles were survey papers that did not describe any AI application, 28 articles were frameworks or tutorials, and 34 articles were papers that did not use any XAI methods, but only mentioned the term \textit{explainable} in the paper without being used in an application. Finally, 187 articles qualified for final inclusion.

\section{Results and Discussion}
\begin{figure}
  \centering
   \begin{tikzpicture}[scale=0.9]
    
%\pie[text=label,style=drop shadow,rotate=180] {10.16/A, 8.56/B, 63.63/C, 17.65/D}%
 \pie[
 /tikz/every pin/.style={align=left},
 text=pin,
       rotate=95,
    color = {
       yellow!60, 
        orange!60, 
       green!30, 
       teal!20},
       explode = 0.2
]
{10/SHAP and LIME \\(Evaluated),
 9/New XAI method \\(Evaluated), 64/SHAP and LIME \\(Not \\evaluated), 17/New XAI \\method \\(Not \\evaluated)}

\end{tikzpicture}
 \caption{Pie chart showing a breakdown of evaluated and not evaluated XAI methods in reviewed papers}
    \label{fig:pie-chart1}
\end{figure}
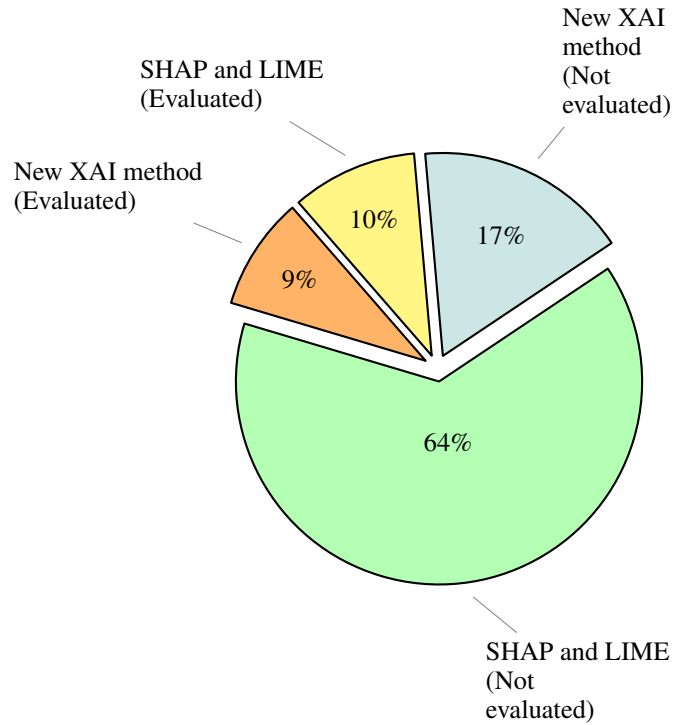

The resultant 187 articles were categorized either as evaluated or not evaluated. There were 35 application papers (19\%) that evaluated the XAI methods and 152 application papers (81\%) that did not evaluate them. Figure~\ref{fig:pie-chart1} shows a pie chart representing the percentage of articles that evaluated the adopted XAI method vs. articles that did not conduct any evaluation. 81\% represents significant level of papers claiming to describe explainable models that do not evaluate the generated explanations. This number is even more concerning when we consider that a large number of those applications may be for high-stake decisions. 

Within the 152 papers that did not evaluate the output of XAI methods, 33 articles included novel approaches. 14 articles utilized LIME libraries and 92 articles adopted SHAP. Finally, 13 articles included a combination of different libraries. Out of 187 application papers, 64\% of the papers used SHAP and LIME feature attribution methods without any evaluation. %Figure~\ref{fig:pie-chart2} shows a pie chart representing the percentage of application papers with a breakdown of evaluated and not evaluated XAI methods. 

Within the category of evaluated XAI methods, the 35 papers found can be further categorized as follows. 16 articles proposed new evaluation methods. This might suggest that these articles were not merely evaluating the explanations of their application but they may have written the paper mostly for introducing the novel method. Only three articles that evaluated explanations had implemented LIME. 10 papers that evaluated their explanations had adopted SHAP. Finally, six articles used the evaluation step to compare the explanations between different XAI methods (\textit{e.g.}, SHAP versus LIME).

Among the 16 articles that used a new evaluation method for their XAI approach, nine articles generated explanations using feature attribution methods, five utilized counterfactual explanations, one used a combination of counterfactual and example-based explanations, and one article utilized an XAI method to generate instance-based explanation. When we compared the evaluation approach of their proposed XAI methods, we found that five articles evaluated the stability and faithfulness of the explanations produced by their model using some kind of metric. For example, Amich and Eshete introduced a novel evaluation metric for stability to evaluate the explanations of their model, EG-Booster. This stability metric enabled vetting the reliability of ML explanation methods before they were used to guide ML robustness evaluation \cite{amich2021egbooster}. For this, they computed the similarity between the returned predictions of the same adversarial sample after EG-Booster attack using (k-l)-Stability and k-Stability metrics. Another novel approach to evaluate the explanation was introduced by Hatwell et al. for their algorithm of the black-box model AdaBoost. They measured the precision, coverage, and stability to evaluate the quality of the explanations generated by their model. To assess the stability and avoid over-fitting explanations, they used the formulation of (n+1)/(m+K), where n is the number of covered and correct instances, m is the number of covered instances, and K is the number of classes \cite{hatwell}. Additionally, they assessed the coverage by evaluating the fraction of previously unseen instances a user could attempt to classify after seeing an explanation and precision by evaluating the fraction of those classifications that would be correct if a user applied the explanation correctly. Böhle et al. took a different approach to evaluate the faithfulness of their model by utilizing a grid-pointing game \cite{bohle2022bcos}. They evaluated the explanations generated by their model on a synthetic 3x3 grid of images of different classes and measured how much positive attribution an explanation method assigned to the correct location in the grid for each of the corresponding class logits. 

Aside from evaluating the stability and faithfulness of the explanations, four articles used different versions of the fidelity metric to evaluate the explanations generated by their models. Panigutti et al. adopted the fidelity metric, Hit, and Explanation complexity to evaluate the different explanation pipelines of their model. As a way to calculate the fidelity to the black box $\in$ [0,1], they compared the predictions made by the interpretable model with the predictions made by the black-box model on a synthetic neighbourhood of the instance and measured the fidelity with micro-averaging  \cite{Panigutti}. Similarly, Setzu et al. used fidelity $\in$ [0,1] to compare the predictions returned by the rules in a given explanation theory E or by the black box b \cite{SETZU2021103457}. Among the XAI methods that generated counterfactual explanations, three articles used different combinations of sparsity, validity, proximity, diversity, distance, and generation time to evaluate the quality of the explanations. For instance, Chen et al. referenced \cite{verma2022counterfactual} to define their evaluation criteria for validity, proximity, sparsity, and generation time of the counterfactual explanations \cite{chen2022relax}. They defined validity as the ratio of the counterfactuals that meet the prediction goal to the total number of the counterfactuals generated, proximity as the distance of a counterfactual from the original input sample, sparsity as the number of features that must be changed according to a counterfactual, and generation time as the time required to generate counterfactuals. Crupi et al. used a similar metric to evaluate their proposed methodology to generate counterfactual explanations, called Counterfactual Explanations as Interventions in Latent Space (CEILS) by using a combination of validity, proximity, sparsity, and distance between the counterfactual and factual observation \cite{crupi2021counterfactual}. 

Although the XAI approaches utilized in the 16 articles were evaluated using some variation of quantitative evaluation, there was an evident lack of qualitative evaluation in these approaches. Since there is still ambiguity in the definition of explanation \cite{gilpin,buchholz2022meansend}, it is significant to place an emphasis on the rigorous qualitative evaluation of user studies \cite{johs2020qualitative} and incorporate usability studies that can improve the user experience \cite{dieber2020model}. Qualitative evaluation methods such as interviews, case studies, focus groups, and observations could be useful in understanding the success and failure of these approaches through a human-centered perspective. Furthermore, most papers that adopted certain evaluation methods did not reference any standard evaluation metrics. Referencing metrics such as \cite{hoffman2019metrics} that include the Explanation Goodness Checklist, Curiosity Check-list, Explanation Satisfaction Scale, and Trust Scale Recommended for XAI, or \cite{metrics_for_evaluating_rosenfeld} that is based on quantifying explainability based on the number of rules in the agent’s explanation, the performance difference between the AI model and the explanation logic, the number of features used to create the explanation, and the stability of the explanation created by the model could be a useful starting point in facilitating a standard practice of evaluating XAI methods. The absence of human-centered and standard evaluation approaches suggest lack of rigour in the field.

\section{Conclusions and Future Work}
Through the scoping review of 187 application papers included in this study, we conclude that most models presented as explainable do not evaluate their XAI methods. This demonstrates a significant gap in publications of AI models that adopt XAI libraries. Furthermore, in the existing literature, evaluation of XAI methods is commonly included when authors propose a new method, or if the study is comparing two or more methods of explanation. We reiterate that XAI methods should be evaluated when they are adopted to produce explanations for a model. Researchers should also be careful when using XAI methods to avoid implying that existing libraries such as SHAP and LIME solve the problem of AI explainability.

In our literature review, we examined the frequency of the practice where papers use the term \textit{explainable model} without evaluating the adopted explanation methods. The main limitation of our study is that it is not an exhaustive review of evaluations of XAI methods which becomes an opportunity for future work. 
 
%% The file named.bst is a bibliography style file for BibTeX 0.99c
\bibliographystyle{named}
\bibliography{ijcai23}

\begin{thebibliography}{}

\bibitem[\protect\citeauthoryear{Adebayo \bgroup \em et al.\egroup
  }{2018}]{adebayo2018sanity}
Julius Adebayo, Justin Gilmer, Michael Muelly, Ian Goodfellow, Moritz Hardt,
  and Been Kim.
\newblock Sanity checks for saliency maps.
\newblock In {\em Advances in Neural Information Processing Systems}, pages
  9505--9515, 2018.

\bibitem[\protect\citeauthoryear{Al-Merri and Ben~Miled}{2022}]{almerri}
Mohammad Al-Merri and Zina Ben~Miled.
\newblock Global translation of classification models.
\newblock {\em Information}, 13(5), 2022.

\bibitem[\protect\citeauthoryear{Alharbi \bgroup \em et al.\egroup
  }{2021}]{alharbi}
Raed Alharbi, Minh~N. Vu, and My~T. Thai.
\newblock Learning interpretation with explainable knowledge distillation.
\newblock {\em CoRR}, abs/2111.06945, 2021.

\bibitem[\protect\citeauthoryear{Alvarez{-}Melis and
  Jaakkola}{2018}]{alavarez2018}
David Alvarez{-}Melis and Tommi~S. Jaakkola.
\newblock Towards robust interpretability with self-explaining neural networks.
\newblock {\em CoRR}, abs/1806.07538, 2018.

\bibitem[\protect\citeauthoryear{Amich and Eshete}{2021}]{amich2021egbooster}
Abderrahmen Amich and Birhanu Eshete.
\newblock Eg-booster: Explanation-guided booster of ml evasion attacks, 2021.

\bibitem[\protect\citeauthoryear{Angelov \bgroup \em et al.\egroup
  }{2021}]{angelov}
Plamen~P. Angelov, Eduardo~A. Soares, Richard Jiang, Nicholas~I. Arnold, and
  Peter~M. Atkinson.
\newblock Explainable artificial intelligence: an analytical review.
\newblock {\em WIREs Data Mining and Knowledge Discovery}, 11(5):e1424, 2021.

\bibitem[\protect\citeauthoryear{Arksey and O'Malley}{2005}]{Arksey}
Hilary Arksey and Lisa O'Malley.
\newblock Scoping studies: towards a methodological framework.
\newblock {\em International Journal of Social Research Methodology},
  8(1):19--32, 2005.

\bibitem[\protect\citeauthoryear{Barda \bgroup \em et al.\egroup
  }{2020}]{barda}
Amie~J. Barda, Christopher~M. Horvat, and Harry Hochheiser.
\newblock A qualitative research framework for the design of user-centered
  displays of explanations for machine learning model predictions in
  healthcare.
\newblock {\em BMC Medical Informatics and Decision Making}, 20(1):257, 2020.

\bibitem[\protect\citeauthoryear{{Barredo Arrieta} \bgroup \em et al.\egroup
  }{2020}]{BARREDOARRIETA202082}
Alejandro {Barredo Arrieta}, Natalia Díaz-Rodríguez, Javier {Del Ser}, Adrien
  Bennetot, Siham Tabik, Alberto Barbado, Salvador Garcia, Sergio Gil-Lopez,
  Daniel Molina, Richard Benjamins, Raja Chatila, and Francisco Herrera.
\newblock Explainable artificial intelligence (xai): Concepts, taxonomies,
  opportunities and challenges toward responsible ai.
\newblock {\em Information Fusion}, 58:82--115, 2020.

\bibitem[\protect\citeauthoryear{Belle and Papantonis}{2021}]{belle}
Vaishak Belle and Ioannis Papantonis.
\newblock Principles and practice of explainable machine learning.
\newblock {\em Frontiers in Big Data}, 4, 2021.

\bibitem[\protect\citeauthoryear{Buchholz}{2022}]{buchholz2022meansend}
Oliver Buchholz.
\newblock A means-end account of explainable artificial intelligence, 2022.

\bibitem[\protect\citeauthoryear{Böhle \bgroup \em et al.\egroup
  }{2022}]{bohle2022bcos}
Moritz Böhle, Mario Fritz, and Bernt Schiele.
\newblock B-cos networks: Alignment is all we need for interpretability, 2022.

\bibitem[\protect\citeauthoryear{Chen \bgroup \em et al.\egroup }{2021}]{hydra}
Yuanyuan Chen, Boyang Li, Han Yu, Pengcheng Wu, and Chunyan Miao.
\newblock Hydra: Hypergradient data relevance analysis for interpreting deep
  neural networks, 02 2021.

\bibitem[\protect\citeauthoryear{Chen \bgroup \em et al.\egroup
  }{2022}]{chen2022relax}
Ziheng Chen, Fabrizio Silvestri, Jia Wang, He~Zhu, Hongshik Ahn, and Gabriele
  Tolomei.
\newblock Relax: Reinforcement learning agent explainer for arbitrary
  predictive models, 2022.

\bibitem[\protect\citeauthoryear{Crupi \bgroup \em et al.\egroup
  }{2021}]{crupi2021counterfactual}
Riccardo Crupi, Alessandro Castelnovo, Daniele Regoli, and Beatriz San~Miguel
  Gonzalez.
\newblock Counterfactual explanations as interventions in latent space, 2021.

\bibitem[\protect\citeauthoryear{Cugny \bgroup \em et al.\egroup
  }{2022}]{cugny}
Robin Cugny, Julien Aligon, Max Chevalier, Geoffrey Roman~Jimenez, and Olivier
  Teste.
\newblock Autoxai: A framework to automatically select the most adapted xai
  solution.
\newblock In {\em Proceedings of the 31st ACM International Conference on
  Information \& Knowledge Management}, CIKM '22, page 315–324, New York, NY,
  USA, 2022. Association for Computing Machinery.

\bibitem[\protect\citeauthoryear{Cui \bgroup \em et al.\egroup
  }{2019}]{cui2019integrative}
Xiaocong Cui, Jung~Min Lee, and J~Hsieh.
\newblock An integrative 3c evaluation framework for explainable artificial
  intelligence.
\newblock In {\em Proceedings of the AI and Semantic Technologies for
  Intelligent Information Systems}, 2019.

\bibitem[\protect\citeauthoryear{Dai \bgroup \em et al.\egroup
  }{2022}]{Dai_2022}
Jessica Dai, Sohini Upadhyay, Ulrich Aivodji, Stephen~H. Bach, and Himabindu
  Lakkaraju.
\newblock Fairness via explanation quality.
\newblock In {\em Proceedings of the 2022 {AAAI}/{ACM} Conference on {AI},
  Ethics, and Society}. {ACM}, jul 2022.

\bibitem[\protect\citeauthoryear{Dieber and Kirrane}{2020}]{dieber2020model}
Jürgen Dieber and Sabrina Kirrane.
\newblock Why model why? assessing the strengths and limitations of lime, 2020.

\bibitem[\protect\citeauthoryear{Doshi-Velez and
  Kim}{2017}]{doshivelez2017rigorous}
Finale Doshi-Velez and Been Kim.
\newblock Towards a rigorous science of interpretable machine learning, 2017.

\bibitem[\protect\citeauthoryear{Gilpin \bgroup \em et al.\egroup
  }{2022}]{gilpin}
Leilani Gilpin, Andrew Paley, Mohammed Alam, Sarah Spurlock, and Kristian
  Hammond.
\newblock ``explanation'' is not a technical term: The problem of ambiguity in
  xai, 06 2022.

\bibitem[\protect\citeauthoryear{GoogleScholar}{2023a}]{GoogleScholar_Shap}
GoogleScholar.
\newblock Google scholar citations for ``a unified approach to interpreting
  model predictions''.
\newblock April 2023.

\bibitem[\protect\citeauthoryear{GoogleScholar}{2023b}]{GoogleScholar_XAI}
GoogleScholar.
\newblock Google scholar citations for titles with the term ``xai'', April
  2023.

\bibitem[\protect\citeauthoryear{GoogleScholar}{2023c}]{GoogleScholar_Influence_functions}
GoogleScholar.
\newblock Google scholar citations for ``understanding black-box predictions
  via influence functions'', April 2023.

\bibitem[\protect\citeauthoryear{GoogleScholar}{2023d}]{GoogleScholar_Lime}
GoogleScholar.
\newblock Google scholar citations for ``why should i trust you?'': Explaining
  the predictions of any classifier'', April 2023.

\bibitem[\protect\citeauthoryear{Gunning \bgroup \em et al.\egroup
  }{2019}]{doi:10.1126/scirobotics.aay7120}
David Gunning, Mark Stefik, Jaesik Choi, Timothy Miller, Simone Stumpf, and
  Guang-Zhong Yang.
\newblock Xai-explainable artificial intelligence.
\newblock {\em Science Robotics}, 4(37):eaay7120, 2019.

\bibitem[\protect\citeauthoryear{Hatwell \bgroup \em et al.\egroup
  }{2021}]{hatwell}
Julian Hatwell, Mohamed~Medhat Gaber, and R.~Muhammad~Atif Azad.
\newblock gbt-hips: Explaining the classifications of gradient boosted tree
  ensembles.
\newblock {\em Applied Sciences}, 11(6), 2021.

\bibitem[\protect\citeauthoryear{Hoffman \bgroup \em et al.\egroup
  }{2017}]{8012316}
Robert~R. Hoffman, Shane~T. Mueller, and Gary Klein.
\newblock Explaining explanation, part 2: Empirical foundations.
\newblock {\em IEEE Intelligent Systems}, 32(4):78--86, 2017.

\bibitem[\protect\citeauthoryear{Hoffman \bgroup \em et al.\egroup
  }{2019}]{hoffman2019metrics}
Robert~R. Hoffman, Shane~T. Mueller, Gary Klein, and Jordan Litman.
\newblock Metrics for explainable ai: Challenges and prospects, 2019.

\bibitem[\protect\citeauthoryear{Huang and
  Marques-Silva}{2023}]{huang2023inadequacy}
Xuanxiang Huang and Joao Marques-Silva.
\newblock The inadequacy of shapley values for explainability.
\newblock {\em arXiv preprint arXiv:2302.08160}, 2023.

\bibitem[\protect\citeauthoryear{Izza \bgroup \em et al.\egroup
  }{2022}]{izza2022tackling}
Yacine Izza, Alexey Ignatiev, and Joao Marques-Silva.
\newblock On tackling explanation redundancy in decision trees.
\newblock {\em Journal of Artificial Intelligence Research}, 75:261--321, 2022.

\bibitem[\protect\citeauthoryear{Johs \bgroup \em et al.\egroup
  }{2020}]{johs2020qualitative}
Adam~J Johs, Denise~E Agosto, and Rosina~O Weber.
\newblock Qualitative investigation in explainable artificial intelligence: A
  bit more insight from social science.
\newblock {\em arXiv preprint arXiv:2011.07130}, 2020.

\bibitem[\protect\citeauthoryear{Keane and Kenny}{2019}]{Keane_2019}
Mark~T. Keane and Eoin~M. Kenny.
\newblock How case-based reasoning explains neural networks: A theoretical
  analysis of {XAI} using post-hoc explanation-by-example from a survey of
  {ANN}-{CBR} twin-systems.
\newblock In {\em Case-Based Reasoning Research and Development}, pages
  155--171. Springer International Publishing, 2019.

\bibitem[\protect\citeauthoryear{Koh and Liang}{2017}]{koh2017understanding}
Pang~Wei Koh and Percy Liang.
\newblock Understanding black-box predictions via influence functions.
\newblock In {\em International conference on machine learning}, pages
  1885--1894. PMLR, 2017.

\bibitem[\protect\citeauthoryear{Lin \bgroup \em et al.\egroup
  }{2020}]{lin2020see}
Yi-Shan Lin, Wen-Chuan Lee, and Z.~Berkay Celik.
\newblock What do you see? evaluation of explainable artificial intelligence
  (xai) interpretability through neural backdoors, 2020.

\bibitem[\protect\citeauthoryear{Liu and Hu}{2022}]{LIU2022102111}
Qian Liu and Pingzhao Hu.
\newblock Extendable and explainable deep learning for pan-cancer radiogenomics
  research.
\newblock {\em Current Opinion in Chemical Biology}, 66:102111, 2022.

\bibitem[\protect\citeauthoryear{Lundberg and Lee}{2017}]{lundberg2017unified}
Scott Lundberg and Su-In Lee.
\newblock A unified approach to interpreting model predictions.
\newblock 2017.

\bibitem[\protect\citeauthoryear{Madumal \bgroup \em et al.\egroup
  }{2021}]{ijcai2021}
Prashan Madumal, Silvia Tulli, Rosina Weber, and David Aha.
\newblock Ijcai workshop on explainable artificial intelligence (xai), 2021.

\bibitem[\protect\citeauthoryear{Marques-Silva and
  Ignatiev}{2022}]{Marques-Silva_Ignatiev_2022}
Joao Marques-Silva and Alexey Ignatiev.
\newblock Delivering trustworthy ai through formal xai.
\newblock {\em Proceedings of the AAAI Conference on Artificial Intelligence},
  36(11):12342--12350, Jun. 2022.

\bibitem[\protect\citeauthoryear{Miller \bgroup \em et al.\egroup
  }{2021}]{ijcai2022}
Tim Miller, Rosina Weber, and Ofra Amir.
\newblock Ijcai 2022 workshop on explainable artificial intelligence (xai),
  2021.

\bibitem[\protect\citeauthoryear{Molnar \bgroup \em et al.\egroup
  }{2020}]{Molnar_2020}
Christoph Molnar, Giuseppe Casalicchio, and Bernd Bischl.
\newblock Interpretable machine learning {\textendash} a brief history,
  state-of-the-art and challenges.
\newblock In {\em {ECML} {PKDD} 2020 Workshops}, pages 417--431. Springer
  International Publishing, 2020.

\bibitem[\protect\citeauthoryear{Montavon}{2019}]{Montavon2019}
Gr{\'e}goire Montavon.
\newblock Gradient-based vs. propagation-based explanations: An axiomatic
  comparison.
\newblock In Wojciech Samek, Gr{\'e}goire Montavon, Andrea Vedaldi, Lars~Kai
  Hansen, and Klaus-Robert M{\"u}ller, editors, {\em Explainable AI:
  Interpreting, Explaining and Visualizing Deep Learning}, pages 253--265.
  Springer International Publishing, Cham, 2019.

\bibitem[\protect\citeauthoryear{Mothilal \bgroup \em et al.\egroup
  }{2020}]{Mothilal}
Ramaravind~K. Mothilal, Amit Sharma, and Chenhao Tan.
\newblock Explaining machine learning classifiers through diverse
  counterfactual explanations.
\newblock In {\em Proceedings of the 2020 Conference on Fairness,
  Accountability, and Transparency}. {ACM}, jan 2020.

\bibitem[\protect\citeauthoryear{Mueller \bgroup \em et al.\egroup
  }{2019}]{mueller2019explanation}
Shane~T. Mueller, Robert~R. Hoffman, William Clancey, Abigail Emrey, and Gary
  Klein.
\newblock Explanation in human-ai systems: A literature meta-review, synopsis
  of key ideas and publications, and bibliography for explainable ai, 2019.

\bibitem[\protect\citeauthoryear{Nugent and Cunningham}{2005}]{cbr-nugent}
Conor Nugent and Padraig Cunningham.
\newblock A case-based explanation system for black-box systems.
\newblock {\em Artif. Intell. Rev.}, 24:163--178, 10 2005.

\bibitem[\protect\citeauthoryear{Page \bgroup \em et al.\egroup
  }{2021}]{prisma}
Matthew~J Page, Joanne~E McKenzie, Patrick~M Bossuyt, Isabelle Boutron, Tammy~C
  Hoffmann, Cynthia~D Mulrow, Larissa Shamseer, Jennifer~M Tetzlaff, Elie~A
  Akl, Sue~E Brennan, Roger Chou, Julie Glanville, Jeremy~M Grimshaw,
  Asbj{\o}rn Hr{\'o}bjartsson, Manoj~M Lalu, Tianjing Li, Elizabeth~W Loder,
  Evan Mayo-Wilson, Steve McDonald, Luke~A McGuinness, Lesley~A Stewart, James
  Thomas, Andrea~C Tricco, Vivian~A Welch, Penny Whiting, and David Moher.
\newblock The prisma 2020 statement: an updated guideline for reporting
  systematic reviews.
\newblock {\em BMJ}, 372, 2021.

\bibitem[\protect\citeauthoryear{Panigutti \bgroup \em et al.\egroup
  }{2020}]{Panigutti}
Cecilia Panigutti, Alan Perotti, and Dino Pedreschi.
\newblock Doctor xai: An ontology-based approach to black-box sequential data
  classification explanations.
\newblock In {\em Proceedings of the 2020 Conference on Fairness,
  Accountability, and Transparency}, FAT* '20, page 629–639, New York, NY,
  USA, 2020. Association for Computing Machinery.

\bibitem[\protect\citeauthoryear{Rafferty \bgroup \em et al.\egroup
  }{2022}]{Rafferty}
Amy Rafferty, Rudolf Nenutil, and Ajitha Rajan.
\newblock Explainable artificial intelligence for breast tumour classification:
  Helpful or harmful.
\newblock In Mauricio Reyes, {Pedro Henriques} Abreu, and Jaime Cardoso,
  editors, {\em Interpretability of Machine Intelligence in Medical Image
  Computing: 5th International Workshop, iMIMIC 2022, Held in Conjunction with
  MICCAI 2022, Singapore, Singapore, September 22, 2022, Proceedings}, Lecture
  Notes in Computer Science, pages 104--123. Springer, Cham, October 2022.

\bibitem[\protect\citeauthoryear{Rawal \bgroup \em et al.\egroup
  }{2022}]{rawal}
Atul Rawal, James McCoy, Danda~B. Rawat, Brian~M. Sadler, and Robert~St. Amant.
\newblock Recent advances in trustworthy explainable artificial intelligence:
  Status, challenges, and perspectives.
\newblock {\em IEEE Transactions on Artificial Intelligence}, 3(6):852--866,
  2022.

\bibitem[\protect\citeauthoryear{Rengasamy \bgroup \em et al.\egroup
  }{2021}]{rengasamy}
Divish Rengasamy, Benjamin~C. Rothwell, and Grazziela~P. Figueredo.
\newblock Towards a more reliable interpretation of machine learning outputs
  for safety-critical systems using feature importance fusion.
\newblock {\em Applied Sciences}, 11(24), 2021.

\bibitem[\protect\citeauthoryear{Ribeiro \bgroup \em et al.\egroup
  }{2016}]{ribeiro2016why}
Marco~Tulio Ribeiro, Sameer Singh, and Carlos Guestrin.
\newblock ``why should i trust you?'': Explaining the predictions of any
  classifier.
\newblock 2016.

\bibitem[\protect\citeauthoryear{Rong \bgroup \em et al.\egroup
  }{2022}]{rong2022humancentered}
Yao Rong, Tobias Leemann, Thai trang Nguyen, Lisa Fiedler, Peizhu Qian, Vaibhav
  Unhelkar, Tina Seidel, Gjergji Kasneci, and Enkelejda Kasneci.
\newblock Towards human-centered explainable ai: User studies for model
  explanations, 2022.

\bibitem[\protect\citeauthoryear{Rosenfeld}{2021}]{metrics_for_evaluating_rosenfeld}
Avi Rosenfeld.
\newblock Better metrics for evaluating explainable artificial intelligence.
\newblock In {\em Proceedings of the 20th International Conference on
  Autonomous Agents and MultiAgent Systems}, AAMAS '21, page 45–50, Richland,
  SC, 2021. International Foundation for Autonomous Agents and Multiagent
  Systems.

\bibitem[\protect\citeauthoryear{Rudin \bgroup \em et al.\egroup
  }{2022}]{rudin2022interpretable}
Cynthia Rudin, Chaofan Chen, Zhi Chen, Haiyang Huang, Lesia Semenova, and Chudi
  Zhong.
\newblock Interpretable machine learning: Fundamental principles and 10 grand
  challenges.
\newblock {\em Statistic Surveys}, 16:1--85, 2022.

\bibitem[\protect\citeauthoryear{Rudin}{2019}]{rudin2019stop}
Cynthia Rudin.
\newblock Stop explaining black box machine learning models for high stakes
  decisions and use interpretable models instead.
\newblock {\em Nature machine intelligence}, 1(5):206--215, 2019.

\bibitem[\protect\citeauthoryear{Setzu \bgroup \em et al.\egroup
  }{2021}]{SETZU2021103457}
Mattia Setzu, Riccardo Guidotti, Anna Monreale, Franco Turini, Dino Pedreschi,
  and Fosca Giannotti.
\newblock Glocalx - from local to global explanations of black box ai models.
\newblock {\em Artificial Intelligence}, 294:103457, 2021.

\bibitem[\protect\citeauthoryear{Shrikumar \bgroup \em et al.\egroup
  }{2019}]{shrikumar}
Avanti Shrikumar, Peyton Greenside, and Anshul Kundaje.
\newblock Learning important features through propagating activation
  differences, 2019.

\bibitem[\protect\citeauthoryear{Slack \bgroup \em et al.\egroup
  }{2019}]{slack}
Dylan Slack, Sophie Hilgard, Emily Jia, Sameer Singh, and Himabindu Lakkaraju.
\newblock How can we fool {LIME} and shap? adversarial attacks on post hoc
  explanation methods.
\newblock {\em CoRR}, abs/1911.02508, 2019.

\bibitem[\protect\citeauthoryear{Sundararajan \bgroup \em et al.\egroup
  }{2017}]{sundararajan2017axiomatic}
Mukund Sundararajan, Ankur Taly, and Qiqi Yan.
\newblock Axiomatic attribution for deep networks.
\newblock {\em arXiv preprint arXiv:1703.01365}, 2017.

\bibitem[\protect\citeauthoryear{Tan \bgroup \em et al.\egroup }{2019}]{tan}
Hui~Fen Tan, Kuangyan Song, Madeilene Udell, Yiming Sun, and Yujia Zhang.
\newblock Why should you trust my interpretation? understanding uncertainty in
  {LIME} predictions.
\newblock {\em CoRR}, abs/1904.12991, 2019.

\bibitem[\protect\citeauthoryear{Tulli \bgroup \em et al.\egroup
  }{2022}]{aaai2022}
Silvia Tulli, Prashan Madumal, Mark~T. Keane, and David~W. Aha.
\newblock Ijcai 2022 workshop on explainable artificial intelligence (xai),
  2022.

\bibitem[\protect\citeauthoryear{{van der Waa} \bgroup \em et al.\egroup
  }{2021}]{VANDERWAA2021103404}
Jasper {van der Waa}, Elisabeth Nieuwburg, Anita Cremers, and Mark Neerincx.
\newblock Evaluating xai: A comparison of rule-based and example-based
  explanations.
\newblock {\em Artificial Intelligence}, 291:103404, 2021.

\bibitem[\protect\citeauthoryear{Verma \bgroup \em et al.\egroup
  }{2022}]{verma2022counterfactual}
Sahil Verma, Varich Boonsanong, Minh Hoang, Keegan~E. Hines, John~P. Dickerson,
  and Chirag Shah.
\newblock Counterfactual explanations and algorithmic recourses for machine
  learning: A review, 2022.

\bibitem[\protect\citeauthoryear{Vu \bgroup \em et al.\egroup }{2019}]{vu2019}
Minh~N. Vu, Truc D.~T. Nguyen, NhatHai Phan, Ralucca Gera, and My~T. Thai.
\newblock Evaluating explainers via perturbation.
\newblock {\em CoRR}, abs/1906.02032, 2019.

\bibitem[\protect\citeauthoryear{Wilming \bgroup \em et al.\egroup
  }{2021}]{wilming2021scrutinizing}
Rick Wilming, Céline Budding, Klaus-Robert Müller, and Stefan Haufe.
\newblock Scrutinizing xai using linear ground-truth data with suppressor
  variables, 2021.

\bibitem[\protect\citeauthoryear{Yang \bgroup \em et al.\egroup
  }{2022}]{yang2022mace}
Wenzhuo Yang, Jia Li, Caiming Xiong, and Steven C.~H. Hoi.
\newblock Mace: An efficient model-agnostic framework for counterfactual
  explanation, 2022.

\bibitem[\protect\citeauthoryear{Yeh \bgroup \em et al.\egroup
  }{2018}]{yeh2018representer}
Chih-Kuan Yeh, Joon~Sik Kim, Ian E.~H. Yen, and Pradeep Ravikumar.
\newblock Representer point selection for explaining deep neural networks,
  2018.

\bibitem[\protect\citeauthoryear{Zhou \bgroup \em et al.\egroup
  }{2021a}]{zhou2021}
Jianlong Zhou, Amir~H. Gandomi, Fang Chen, and Andreas Holzinger.
\newblock Evaluating the quality of machine learning explanations: A survey on
  methods and metrics.
\newblock {\em Electronics}, 10(5), 2021.

\bibitem[\protect\citeauthoryear{Zhou \bgroup \em et al.\egroup
  }{2021b}]{zhou2021feature}
Yilun Zhou, Serena Booth, Marco~Tulio Ribeiro, and Julie Shah.
\newblock Do feature attribution methods correctly attribute features?
\newblock {\em CoRR}, abs/2104.14403, 2021.

\end{thebibliography}

\end{document}